\documentclass[nonatbib]{article}

\usepackage{neurips_2020}

\usepackage[utf8]{inputenc} 
\usepackage[T1]{fontenc}    
\usepackage{xr-hyper}
\usepackage{hyperref}       
\usepackage{url}            
\usepackage{booktabs}       
\usepackage{amsfonts}       
\usepackage{nicefrac}       
\usepackage{microtype}      
\usepackage{mathtools}
\usepackage{color}
\usepackage[square,numbers]{natbib}
\usepackage{caption}
\usepackage{subcaption}
\DeclarePairedDelimiter{\norm}{\lVert}{\rVert}

\usepackage[ruled,vlined]{algorithm2e}
\usepackage{etoolbox}
\hypersetup{
    colorlinks=true,
    linkcolor=blue,
    filecolor=magenta,      
    urlcolor=magenta,
}
\DontPrintSemicolon

\SetKwComment{Comment}{$\triangleright$\ }{}
\usepackage{floatrow}

\newfloatcommand{capbtabbox}{table}[][\FBwidth]
\title{Compression-aware Continual Learning using Singular Value Decomposition}

\author{%
  Varigonda Pavan Teja  \\
University of Tübingen, Germany \\
  \texttt{pavan.teja295@gmail.com} \\
   \And
   Priyadarshini Panda \\
   Department of Electrical Engineering\\
   Yale University\\
   New Haven, CT, 06511\\
   \texttt{priya.panda@yale.edu}
}

\begin{document}

\maketitle
\begin{abstract}
We propose a compression based continual task learning method that can dynamically grow a neural network. Inspired from the recent model compression techniques, we employ compression-aware training and perform low-rank weight approximations using singular value decomposition (SVD) to achieve network compaction. By encouraging the network to learn low-rank weight filters, our method achieves compressed representations with minimal performance degradation without the need for \emph{costly} fine-tuning. Specifically, we decompose the weight filters using SVD and train the network on incremental tasks in its factorized form. Such a factorization allows us to directly impose \emph{sparsity-inducing} regularizers over the singular values and allows us to use fewer number of parameters for each task. We further introduce a novel shared representational space based learning between tasks. This promotes the incoming tasks to only learn residual \emph{task-specific} information on top of the previously learnt weight filters and greatly helps in learning under fixed capacity constraints. Our method significantly outperforms prior continual learning approaches on three benchmark datasets, demonstrating accuracy improvements of \emph{10.3\%}, \emph{12.3\%},  \emph{15.6\%} on 20-split CIFAR-100, miniImageNet and a  5-sequence dataset, respectively, over state-of-the-art. Further, our method yields compressed models that have $\sim 3.64\times, 2.88\times, 5.91\times $ fewer number of parameters respectively, on the above mentioned datasets in comparison to baseline individual task models. Our source code is available at \url{https://github.com/pavanteja295/CACL}.

\end{abstract}

\section{Introduction}
The ability to learn novel information without interfering with the consolidated previous knowledge is referred to as \emph{life-long} or \emph{continual} learning. Deep learning methods have demonstrated superseding performances and improvements on single-task learning \cite {AlexNet} and multi-task learning \cite{zhang2017survey}  that aims to jointly optimize over several tasks at once. Albeit this success, neural networks suffer from \emph{catastrophic forgetting} during sequential-task learning and lack the ability to incrementally learn new information. This is ascribed to the  representational overlap between tasks resulting in interference of new information with previous knowledge \cite{French_CN}.

Recent studies have proposed methods that alleviate the catastrophic forgetting problem. Some approaches referred to as \emph{regularization} based methods estimate the contribution of the individual parameters in a network  to prior tasks and \emph{regularize} the weight changes/updates based on the estimated importance while training new tasks \cite{Aljundi_2018_ECCV, Kirkpatrick2017OvercomingCF, SI_G, Serr2018OvercomingCF}. However, these methods assume fixed network capacity that upper bounds the number of tasks that can be learnt. In fact, the authors in \cite{Hsu18_EvalCL} observed degrading performance of such approaches over longer task sequences. Other line of work categorized as \emph{memory-replay} methods \cite{Chaudhry2019ContinualLW, GEM, Rolnick2018ExperienceRF, Chaudhry2019EfficientLL} retain examples from previous tasks to retrain the network that helps restore performances on old tasks. But these methods are burdened with \emph{retraining} on all previous tasks for every new task.

\emph{Network expansion} methods consider growing the network to avoid the representational overlap between tasks and accommodate unlimited number of tasks. Further, some of these approaches \cite{Rusu2016ProgressiveNN, Mallya2018PackNetAM,PiggyBack, hung2019compacting} guarantee preservation of performances on previous tasks and eliminate retraining on old tasks. Nevertheless, unrestricted growing of networks is computationally expensive and memory intensive that hinders their real-time implementation on hardware with limited memory. 
Approaches like \cite{PAE, Mallya2018PackNetAM, hung2019compacting, saha2020structured} employ pruning procedures to achieve network compaction, thereby, reducing the memory consumption. However, such pruning procedures are either iterative and gradual or require an additional fine-tuning/retraining step to restore the performance. Other works \cite{yoon2018lifelong, hung2019compacting} control the model growth by introducing \emph{carefully} designed expansion schemes that are \emph{time-consuming}.

In this paper, we propose a network expansion method for continual learning 
addressing the above mentioned weaknesses while demonstrating \emph{significant} improvements to average performance and model compaction. By adopting two main characteristics: 1) \emph{Network Factorization}, and 2) \emph{Additive Shared Representational Space}, we achieve \emph{state-of-the results} with \emph{best} model compression rates on three benchmark datasets. Our compression step produces compact representations in \emph{oneshot} for each task with minimal performance degradation unlike the \emph{iterative} and \emph{time-consuming}  pruning strategies used in prior works. Our compression technique is inspired from the recent \emph{low-rank} approximation methods \cite{Denton2014ExploitingLS, Comp_aware_training, saha2020structured} using SVD that compress networks via singular value pruning. We resort to compression-aware training by encouraging the model to learn low-rank weight filters by enforcing sparsity regularizers over singular values similar to \cite{yang2020learning,Comp_aware_training, Xu2018TrainedRP}. This minimizes the performance degradation during the compression step eliminating the costly \emph{retraining}. Moreover, we achieve maximal compression with a novel \emph{additive} shared representation learning between tasks. Our proposed sharing technique enables adding residual task-specific information pertaining to the incoming tasks on the already learnt weight filters from prior tasks. We employ a simple unconstrained expansion scheme to accommodate new incoming tasks.

To summarize, our contributions are:
\begin{itemize}
\itemsep-0.2em 
    \item We propose a simple yet effective incremental task learning algorithm that does not resort to \emph{time-consuming} intermediate heuristics.
    \item We employ compression-aware training of neural networks in the SVD decomposed form to tackle sequential task learning. We compress the trained task representations using low-rank approximations and incorporate incoming tasks by expansion.
    \item We propose \emph{additive} shared representations between tasks in the factorized space that encourages the incoming tasks to learn \emph{task-specific} information.
    \item We show significant improvements in performance over \emph{state-of-the-art} approaches on benchmark datasets, CIFAR100 \cite{CIFAR100},  miniImageNet \cite{ miniImageNet} and 5-sequence dataset \cite{ebrahimi2020adversarial} with much \emph{smaller} model size. Through a fair comparison with other network-based expansion methods, we show that our approach is highly optimal by \emph{remarkably} outperforming in accuracy and compression.

\end{itemize}

\section{Related Work}
\label{r_work}

\textbf{Regularization based methods}  control the updates to the weights based on their degree of importance per task. 
EWC \cite{Kirkpatrick2017OvercomingCF} identifies significant weights by using Fischer Information Matrix, while authors in \cite{SI_G} estimate importance of parameters by their individual contribution over the entire loss trajectory. Improvements to efficiency and memory consumption of  EWC was proposed in \cite{OnlineEWC}.
HAT \cite{Serr2018OvercomingCF} proposes to learn attention mask to control the weight updates while training on new tasks. The authors in \cite{Uncertainty} employ a Bayesian framework where, parameters predicted with high certainty are considered crucial to maintain previous task performance.

\textbf{Memory replay methods} deal with the catastrophic forgetting of previous experiences by rehearsing the old tasks while training the new tasks. This is achieved by either storing subset of data \cite{GEM, chaudhry2018efficient, NIPS2019_8327, Chaudhry2019EfficientLL, Chaudhry2019ContinualLW} from previous tasks or by synthesizing the old data using generative models \cite{kemker2018fearnet, NIPS2017_6892}. Sample selection strategy plays an important role in these approaches due to the limited computational budget. Significant improvements  in performance were demonstrated by \cite{Chaudhry2019ContinualLW, L2lwfbmt} by employing reservoir sampling than random sampling for selecting examples from old tasks.

\textbf{Network expansion methods} increases the network capacity to prevent interference among task representations during incremental task learning. Our work largely belongs to this line of work.   ProgressiveNet \cite{Rusu2016ProgressiveNN}  accommodates new tasks by instantiating a new network for each task  but at the cost of linear architectural growth. To avoid this, the authors in \cite{yoon2018lifelong} allow network expansion only when required based on the task relatedness to the previous knowledge. PiggyBack \cite{PiggyBack} learns a selection mask per task to adapt a fixed backbone network to multiple tasks and requires a  \emph{pre-trained} backbone. In contrast to PiggyBack, our method can continually learn from scratch.
Our approach is closely related to \cite{hung2019compacting, PAE} that follow a \emph{learn-compress-grow} cycle for each task. However, the above methods use gradual pruning and compression schemes that are time-consuming. 
Another recent work \cite{saha2020structured} also focuses on compression for continual learning tasks. The authors in \cite{saha2020structured} use Principal Component Analysis\cite{PCA} on the feature maps to detect the relevant task-specific filters and remove redundant filters while learning tasks sequentially. However, they perform compression across channels of different layers and do not preserve the initial network design unlike our approach. Furthermore, the authors in \cite{saha2020structured} train the network in a generic manner and then use PCA to determine the optimal task-specific filters to prune/retain. Then, they employ a fine-tuning/re-training step after every compression step to recover the performance degradation due to compression. In contrast to \cite{saha2020structured}, the novelty of our method is that we train the network in the SVD factorized space that encourages low-rank approximation and sparsification. This SVD space continual training provides us with the advantage of eliminating the re-training step while achieving maximal compression without any significant performance loss.




\textbf{Low rank approximation}:  Recent works \cite{Denton2014ExploitingLS, Jaderberg2014SpeedingUC} demonstrated significant model compression rates using low-rank approximations of the weights of a network. To minimize loss in accuracy due to the compression, the authors in \cite{Comp_aware_training, Xu2018TrainedRP, yang2020learning} propose embedding low-rank decomposition during training by imposing sparsity over the singular values to learn low-rank weight filters. 

\section{Compression-aware Continual Learning  (CACL)}
\begin{figure}
    \centering
    \vspace{-3em}
    \includegraphics[width=\linewidth]{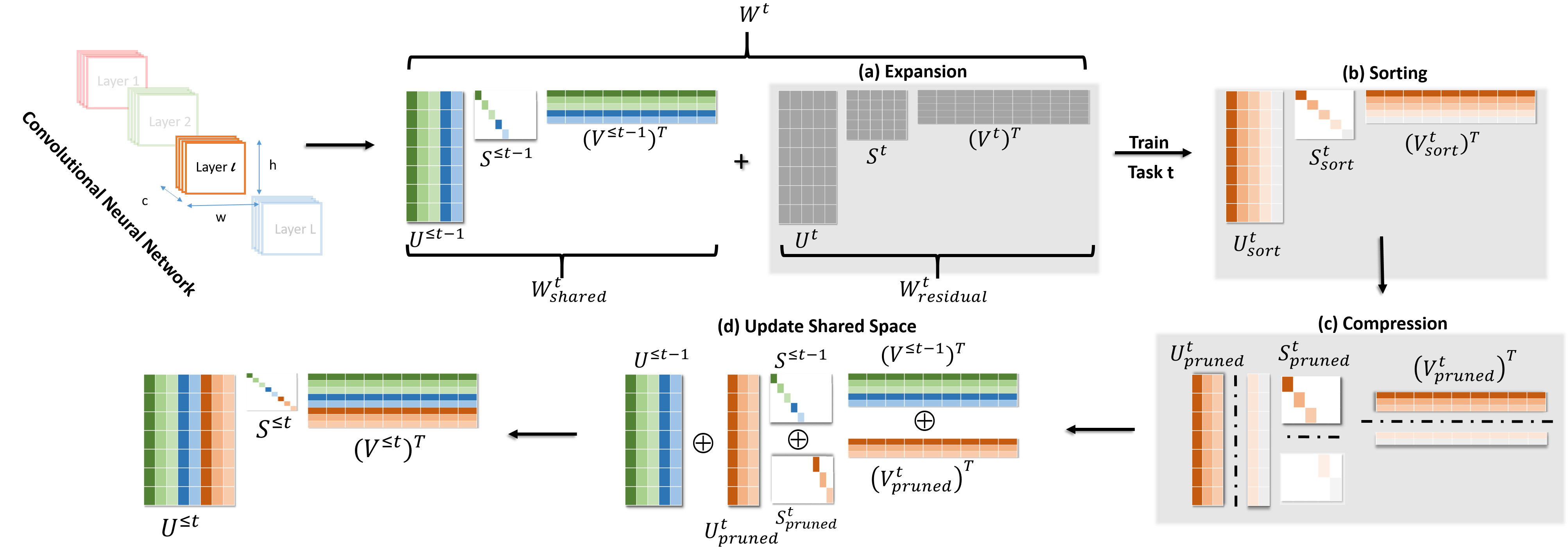} 

    \captionof{figure}{A schematic of CACL training pipeline for a layer $l$ and task $t$ is shown. For simplicity, we drop the notation `$l$' from all the parameters. (a) We start with \emph{expansion} by creating randomly initialized trainable parameters $U^t, V^t, S^t$. We then train the parameters $U^t, V^t, S^t$ following Section \ref{training}. (b) We obtain the sorted singular values and vectors $U^t_{sort}, V^t_{sort}, S^t_{sort}$ by performing \emph{sorting} on the singular values after training. (c) We apply our \emph{compression} step as discussed in Section \ref{compression} that results in $U^t_{pruned}, V^t_{pruned}, S^t_{pruned}$. (d) Finally, the compressed singular values and vectors are \emph{added} to shared space that returns the $U^{\leq t}, V^{\leq t}, S^{\leq t}_l$ used by the next incoming task \emph{t+1}.}
     \label{fig:pipeline}
\end{figure}

We consider the problem of incremental task learning where, $T$ tasks with unknown data-distributions are to be learnt in a \emph{sequence}  with the help of training data 
$\mathcal{D}_t$ 
where $\mathcal{D}_t$ = $\{x_i, y_i\}_{i=1}^{N_{t}}$ and $t \in \{1, 2, ... T\}$. In the sections below, we discuss the steps for our proposed CACL methodology. An overview of CACL is outlined in Fig. \ref{fig:pipeline} and Algorithm \ref{train_algorithm}. In essence, for any incoming task $t$, we instantiate a convolutional network in SVD parameterized form (Section \ref{expansion}). We train the factorized network by incorporating sharing between tasks (Section \ref{sec_sharing}) and impose sparsity-inducing regularizer to encourage low-rank weight solutions (Section \ref{training}). We finally compress the task representations using low rank approximations of the weight filters and update the shared space by appending the compressed representations (Section \ref{compression}).

\subsection{Background}  
\label{background_ours}
\textbf{Singular Value Decomposition}: SVD factorization of any \emph{real} 2-dimensional matrix $\mathcal{A}$ $\in$ $R^{m \times n}$ is given by $U diag(\sigma) V^T$ where $U$ $\in R^{m \times m}$,  $V \in R^{n \times n}$, $diag(\sigma)$  $\in$  $R^{m \times n}$ are left singular vectors, right singular vectors, diagonal matrix with singular values on the principal diagonal, respectively.  By construction, $U^TU = UU^T = I$ ,  $V^TV = VV^T = I$.
One could consider a \emph{reduced} SVD form by only retaining the non-zero singular values and their corresponding singular vector columns. Then $\mathcal{A}$ can be safely factorized into $U$ $\in R^{m \times r}$,  $V \in R^{n \times r}$, $diag(\sigma)$  $\in$  $R^{r \times r}$. Dimension $r$ refers to the rank of the matrix which is the number of non-zero singular values. We refer to \emph{reduced} SVD as SVD in the subsequent sections.


\textbf{Rank-k approximation}:
For any real matrix $\mathcal{A}$ $\in$ $R^{m \times n}$  and $m \geq n$, the \emph{best} rank-k approximation is given by 
$
\mathcal{A}_{k} = \sum_{i=1}^{k} \sigma_{i}u_{i}v_{i}^T$ where, $\sigma_{1} \geq \sigma_{2} ... \geq \sigma_{n} \geq 0 $ are the \emph{sorted} singular values. Here, $u_{i} \in$  $R^{m \times 1}$, $v_{i}$ $\in$  $R^{n \times 1}$ denote the corresponding left and right singular columns from $U$ and $V$, respectively. Rank-k approximation using SVD essentially prunes the insignificant singular values ($k+1$ to $n$) and their corresponding singular vectors which ultimately enables compression in our approach. However, such an approximation introduces error. The \emph{Frobenius norm} due to the substitution of $\mathcal{A}$ with $\mathcal{A}_{k}$ is given as
$
\norm{ \mathcal{A} - \mathcal{A}_{k}}_{F}  = \sum_{i=k+1}^{n} \sigma^{2}_{i}$ where, $\norm{.}_F$ represents the Frobenius norm. The difference between $\mathcal{A}$ and $\mathcal{A}_{k}$ is highly minimized when $\mathcal{A}$ is sufficiently low-rank. Thus, we encourage the model to learn low-rank weight filters during training by imposing sparsity-inducing regularizer over singular values.


\textbf{Notations}: For convenience, we provide the reader the notations used in the subsequent sections and their definitions. $\mathcal{W}^t = \{ W^t_l \}^L_{l=1}$ represents the weight tensors of a neural network for task $t$ across all layers where, $W^t_l$ corresponds to the weight tensor of task $t$ at layer $l$. Similarly, we denote $\mathcal{U}^t = \{U^t_l\}_{l=1}^L $,    $\mathcal{V}^t = \{V^t_l\}_{l=1}^L$, $\mathcal{S}^t = \{S^t_l\}_{l=1}^L = \{diag(\sigma)^t_l\}_{l=1}^L $. $U^t_l$, $V^t_l$ denotes the left and right singular vectors, while $S^t_l$ or $diag(\sigma)^t_l$ denotes the diagonal matrix with singular values on the principal diagonal. $\sigma^t_l$ denotes the singular values in vectorized form. We consider $\mathcal{U}^t, \mathcal{V}^t, \mathcal{S}^t$ to be the left singular vectors,  right singular vectors, singular values, respectively, across all layers for task $t$. Any operation on $\mathcal{U}^t, \mathcal{V}^t, \mathcal{S}^t$ unless otherwise specified is meant to be applied for each layer separately.

\subsection{Network Factorization}
 \label{factorization}
    Our motivation to train and store the learnt representations of a neural network in the factorized SVD form is drawn from two advantages. Firstly, it allows preserving the \emph{low-rank} weight filters  with fewer number of parameters. For a convolutional layer $W_{conv} \in R^{c \times n \times h \times w}$ where $c$ is the number of output channels, $h \times w$ is size of the kernel, $n$ is the number of input channels, the number of parameters required to store a 4D weight matrix is \emph{cnhw}. However, when $W_{conv}$ is sufficiently low-rank, this representation could be sub-optimal. Thus, we reshape the 4D tensor to a 2D matrix and decompose the matrix using SVD  into $U_{conv}$ $\in$ $R^{c \times r}$, $V_{conv}$ $\in$ $R^{nhw \times r}$, $diag(\sigma)_{conv}$ $\in$ $R^{r \times r}$.
    The SVD parameterized form requires \emph{cr + nhwr + r}  parameters. For $r << \frac{cnhw}{c + nhw + 1}$, 
    the factorized representation \emph{significantly} reduces the memory consumption compared to original representation.
    Secondly, one can directly impose sparsity regularizers over the singular values to promote learning low-rank filters, without resorting to costly SVD factorization at every training step.

    We maintain the factorized form throughout the \emph{training} as well as \emph{inference} that adds a small computational overhead to construct the 4D convolutional weight filters from the decomposed parameters. As a result for any task $t$, our trainable parameters are $\mathcal{U}^t, \mathcal{V}^t, \mathcal{S}^t$. Note that, such a factorization does not mutate the architecture of the network, but rather modifies the representations we learn and store.

\subsection{Shared Representational Space}
\label{sec_sharing}
For any incoming task $t$, we reuse the previously accumulated experiences from old tasks and promote tasks to learn on top of the previous weight filters. This encourages learning task-specific \emph{residual} information.
We introduce a novel sharing scheme in the factorized space that allows incoming tasks to leverage previous knowledge. This is enabled by 
constructing the \emph{2D reshaped} (reshaping 4D tensor to 2D matrix as explained in Section \ref{factorization}) convolutional weight filters while training a task  $t$ as below: 
\begin{equation}
\label{eq:sharedrepresentation}
 \begin{aligned}
 \mathcal{W}^{t}  =  \underset{\text{Frozen}}{\boxed{\mathcal{U}^{\leq t-1}\mathcal{S}^{\leq t-1}(\mathcal{V}^{\leq t-1})^T}} +
 \underset{\text{Trainable}}{\boxed{\mathcal{U}^{t}\mathcal{S}^{t}(\mathcal{V}^{t})^T}}
 \end{aligned}
\end{equation}
where $+$ represents element-wise addition  across all layers. 
As shown in Eqn. \ref{eq:sharedrepresentation}, during task $t$ training, the trainable parameters are $\mathcal{U}^{t}, \mathcal{S}^{t}, \mathcal{V}^{t}$, while $\mathcal{U}^{\leq t-1}, \mathcal{S}^{\leq t-1}, \mathcal{V}^{\leq t-1}$ are frozen. Intuitively, one can interpret our sharing mechanism as \emph{adding} residual task related weight filters ($\mathcal{W}^t_{residual}$) to the previously learnt weights ($\mathcal{W}^t_{shared}$)
where, $\mathcal{W}^t_{residual} = \mathcal{U}^{t} \mathcal{S}^{t} (\mathcal{V}^{t})^T$ and $\mathcal{W}^t_{shared} = \mathcal{U}^{\leq t-1} \mathcal{S}^{\leq t-1} (\mathcal{V}^{\leq t-1})^T$. Note, our sharing technique maintains the intended network design (number of layers or number of weight filters within each layer) unlike previous network expansion approaches \cite{yoon2018lifelong, hung2019compacting}. 

    \begin{minipage}[t]{\textwidth}
\begin{algorithm}[H]
\small
\SetAlgoLined
\SetKwInOut{Input}{Input}
\Input{Dataset $\mathcal{D}$ = ($D_1, D_2, ... D_T$), $\mathcal{W} = \{W_l\}_{l=1}^L$, $Threshold = e$ }
\;
$\mathcal{U}^{t=0}, \mathcal{S}^{t=0}, \mathcal{V}^{t=0} = [\;]  $\Comment*[r]{Intialization } 

 \For{$t\gets1, 2, ... T$}{
    $\mathcal{U}^{t}, \mathcal{S}^{t}, \mathcal{V}^{t} = \; $\emph{Expansion} ($\mathcal{W}$) \Comment*[r]{Section \ref{expansion} }
    $\mathcal{U}^{t}, \mathcal{S}^{t},
    \mathcal{V}^{t} =  \emph{TrainTask}( D_{t}$,  $[\mathcal{U}^{\leq t-1}, \mathcal{S}^{\leq t-1}, \mathcal{V}^{\leq t-1} ]$,
    $[\mathcal{U}^{t}, \mathcal{S}^{t}, \mathcal{V}^{t}]$) \Comment*[r]{Section \ref{training} }
    $\mathcal{U}^{t}_{pruned}, \mathcal{S}^{t}_{pruned}, \mathcal{V}^{t}_{pruned} = \; $\emph{Compression} ( $[\mathcal{U}^{t}, \mathcal{S}^{t}, \mathcal{V}^{t}]$, $e$)  \Comment*[r]{Section \ref{compression} }
    $\mathcal{U}^{\leq t}, \mathcal{S}^{\leq t}, \mathcal{V}^{\leq t} = \;$ $\mathcal{U}^{\leq t-1} \bigoplus \mathcal{U}^{t}_{pruned}, \; \mathcal{S}^{\leq t-1} \bigoplus \mathcal{S}^{t}_{pruned} , \; \mathcal{V}^{\leq t-1} \bigoplus \mathcal{V}^{t}_{pruned} $
    
    }
 \caption{Compression-aware Continual Learning (CACL)}
 \label{CACL_algorithm}
\end{algorithm}
\end{minipage}



\subsection{Expansion}
\label{expansion}
	For any incoming task $t$, we create randomly initialized parameters namely, $\mathcal{U}^{t}, \mathcal{S}^{t}, \mathcal{V}^{t}$. Specifically, for each layer $l$  of a given network, we create \emph{trainable} parameters $U^{t}_l \in R^{c \times r}, V^t_l \in R^{nhw \times r}, S^{t}_l \in R^{r \times r}$ that learns the task-specific residual information (See Fig. \ref{fig:pipeline}(a)). The dimensions $c, n, h, w$ denote the output channels, input channels, kernel height and kernel width of the weight filter in layer $l$ of the given network, respectively. The dimension $r$  of the instantiated matrices is equal to $r = \frac{cnhw}{c + nhw + 1} $ to ensure the number of trainable parameters in the factorized network be same as the original network. We further create an individual task-head for task $t$ referred to as $T^t_{head}$ attached on top of the network that predicts the final task-specific output. 

\subsection{Compression-Aware Training} 
\label{training}
We learn task $t$ by training the parameters $\mathcal{U}^t, \mathcal{S}^t, \mathcal{V}^t, T^t_{head}$ created during the expansion step using the task $t$ dataset $D_t = \{x_i, y_i\}_{i=1}^{N_t}$. 
	The convolutional weight filters ($\mathcal{W}^t = \mathcal{W}^t_{shared} + \mathcal{W}^t_{residual}$) at every training step are constructed using the sharing technique (Eqn. 1) discussed in Section \ref{sec_sharing}. Task $t$ is learnt by applying the task-specific loss function denoted as $\mathcal{L}_{task}$ on the task-head $T_{head}^t$. Throughout the training, we ensure $\mathcal{W}^t_{residual}$  maintains a valid SVD form so that we can
obtain the \emph{best} low-rank approximations during our compression step (see Section \ref{background_ours}). For this, we require parameters $\mathcal{U}^t, \mathcal{V}^t$ to be orthogonal i.e. $\mathcal{U{U}}^T = \mathcal{U}^T\mathcal{U} = I$ and $\mathcal{V{V}}^T =  \mathcal{V}^T\mathcal{V} = I$. Hence,  we employ an \textbf{orthogonality-regularizer} ($\mathcal{L}_{orth}$) proposed in  \cite{yang2020learning} on parameters $\mathcal{U}^t, \mathcal{V}^t$  which is given by: 
$\mathcal{L}_{orth}  = \sum_{l=1}^L \dfrac{1}{r^2}(\norm{(U^t_l)^TU^t_{l} - I}_F + \norm{(V^t_l)^TV^t_{l} - I}_F)$ where, $\norm{.}_F$ is the Frobenius norm, $I$ is the identity matrix, $r$ refers to the \emph{rank} of the parameters $U^t_l$ and $V^t_l$.




Note, low-rank approximations of the weight filters could introduce errors that can cause severe performance degradation during the compression phase (See Section\ref{background_ours}). This degradation can be minimized by learning low rank weight filters during training. To accomplish this, we impose \emph{sparsity-inducing} regularizer over the singular values during training. We employ \emph{Hoyer} regularization \cite{yang2020learning} that was observed to achieve better singular value sparsity than others (such as $L_1$ and $L_2$ norms). We denote the \emph{Hoyer} regularization term as $\mathcal{L}_{sparse}$ where, $\mathcal{L}_{sparse}=\sum_{l=1}^L \dfrac{\norm{\sigma^t_l}_1}{\norm{\sigma^t_l}_2}$. $\norm{.}_1$ denotes the $L_1$ norm and $\norm{.}_2$ refers to the $L_2$ norm.

The overall loss function and the optimization objective while training a task $t$ corresponds to:
\begin{gather}
\begin{aligned}
\mathcal{L}_{total}  = \mathcal{L}_{task} +  \lambda_{orth}\mathcal{L}_{orth} + \lambda_{sparse}\mathcal{L}_{sparse} \\ 
\underset{\mathcal{U}^t, \mathcal{V}^t, \mathcal{S}^t}{\text{minimize}} \quad 
\mathcal{L}_{total}(  \mathcal{U}^t, \mathcal{V}^t, \mathcal{S}^t \textbf{;}\quad                                \mathcal{U}^{\leq t-1},\mathcal{S}^{\leq t -1},\mathcal{V}^{\leq t-1}, \mathcal{D}_t)
\end{aligned}
\end{gather}
where $\lambda_{orth}, \lambda_{sparse}$ are the hyper-parameters to control the effect of each component and  $D_t$ refers to the training data of task $t$. We provide the pseudo code of task training ($TrainTask$ function in Algorithm 1) in the supplementary material.

\subsection{Compression}
\label{compression}

We obtain trained $\mathcal{U}^t, \mathcal{S}^t, \mathcal{V}^t$ parameters and perform singular value pruning at each layer \emph{l}  to achieve low-rank approximations of $\mathcal{W}_{residual}$. We initially sort the singular values in $\mathcal{S}^t$ and their corresponding singular vectors in descending order which returns $\mathcal{U}_{sort}^t, {S}_{sort}^t, {V}_{sort}^t$ (See Fig. \ref{fig:pipeline}(b)). We then retain only the  top \emph{k} highest singular values and their corresponding singular vectors eliminating the insignificant singular values. This results in the \emph{rank-k} approximation of $\mathcal{W}_{residual}$ and outputs $\mathcal{U}_{pruned}^t {S}_{pruned}^t, {V}_{pruned}^t$ with the rank of the parameters equal to $k$ (See Fig. \ref{fig:pipeline}c). The top $k$ value determines the number of singular values to retain. However, resorting to a constant $k$ for all the tasks and layers without task-consideration will be sub-optimal. Hence, we follow the below heuristic inspired from \cite{Xu2018TrainedRP}. We prune the singular values based on total singular value energy which is given by, $\sum_{i=k+1}^n (\sigma_{i})^2 \leq e \sum_{j=1}^k (\sigma_{j})^2$,
where $e$ is a hyperparameter that controls the pruning intensity and $\sigma_{i}, \sigma_{j} \in \sigma^t_l$. The above heuristic allows \emph{dynamic} memory allocation to different tasks, that is a favourable characteristic in incremental task learning approaches. We provide a psuedo-code of our compression algorithm in the supplementary material.

Finally, compressed task representations $\mathcal{U}_{pruned}^t, \mathcal{S}_{pruned}^t, \mathcal{V}_{pruned}^t$ are appended to the existing shared space $\mathcal{U}^{\leq t-1}, \mathcal{S}^{\leq t-1}, \mathcal{V}^{\leq t-1}$ that results in a new shared space $\mathcal{U}^{\leq t}, \mathcal{S}^{\leq t}, \mathcal{V}^{\leq t}$. We store the rank of the parameters in the new shared space as \emph{task-identifiers} for task $t$. During \emph{inference}, we use the task-identifier of a task $t$ to extract the sub network $\mathcal{U}^{\leq t}, \mathcal{S}^{\leq t}, \mathcal{V}^{\leq t}$. For a $L$ layered neural network and $T$ tasks, storing the task-identifier incurs an overhead of \emph{$L\times T$} integer values.

\section{Experiments}
\label{experiments}
We evaluate our approach on three benchmark datasets - CIFAR-100\cite{CIFAR100}, miniImageNet\cite{miniImageNet}, 5-sequence dataset \cite{ebrahimi2020adversarial}. We follow the evaluation protocols proposed by recent works \cite{ebrahimi2020adversarial, hung2019compacting} for a fair comparison with other approaches. We use a simple 5 layer convolutional neural network to demonstrate the effectiveness of our approach. We use a single hidden layer followed by softmax for each task as the task-head (${T^t}_{head}$)
to output task specific classification scores. Unless explicitly mentioned, we use the above architecture for all our experiments. We refer the reader to supplementary section for the architectural details and hyperparameter configurations. Although our SVD factorization and compression-aware training approach can be extended to linear layers, in our experiments we apply it on \emph{convolutional} layers. Our approach is referred as $CACL$ in the results.

We further train and evaluate a simple \emph{baseline} that serves as the \emph{upper bound} on the average performance for our approach. The baseline consists of individual single task models trained separately for each task using the above mentioned 5 layer convolutional network. We refer to this as \emph{Baseline\textunderscore UB}.
All the results reported on our approach are averaged across 3 runs with random task ordering and weight initialization as prior works. We would further like to note that the overhead observed for training the 5 layer convolutional network in factorized form using CACL is marginal compared to standard training.

\textbf{Evaluation metrics}: We report the test classification accuracy averaged across all tasks abbreviated as \emph{ACC}. We further report the size of the final trained models in Mega Bytes(assuming 32-bit floating point is equivalent to 4 bytes). We also report the backward transfer (\emph{BWT}) introduced in \cite{GEM}, that denotes the average forgetting and the influence of learning new tasks on previous tasks. Lower \emph{BWT} score implies better continual learning.

\subsection{20-split CIFAR-100}
\label{cif-100}
We conduct two experiments on CIFAR-100 dataset, referred to as \textbf{Protocol 1} and \textbf{Protocol 2}. For both protocols, we split CIFAR-100 dataset \cite{CIFAR100} into 20 disjoint tasks learning 5 class classification per task.  We discuss below the motivation and the results obtained on each protocol using our method. 

\textbf{Protocol 1}: In this protocol, we follow evaluation settings proposed by the recent continual learning methods \cite{ebrahimi2020adversarial, Chaudhry2019EfficientLL}. The focus of this experiment is to achieve the best performance and show the effectiveness of our CACL approach using a simple 5-layer convolutional network (mentioned above) without any additional components(such as residual connections \cite{ResNet}, batch normalization \cite{BatchNorm}). 
Note, the results of the recent works reported in Table \ref{protocol1} have been taken from \cite{ ebrahimi2020adversarial}. Our best model with sparsity loss $\lambda_{sparse}=0.1$ \emph{significantly} surpasses the previous approaches in average accuracy while maintaining minimal memory consumption. Our method shows \textbf{10.3\%} improvement in average accuracy using $\sim$\textbf{2.89}$\times$ fewer number of parameters when compared to the \emph{state of the art} Adversarial Continual Learning (ACL) \cite{ebrahimi2020adversarial}. We also report another CACL model trained with a higher sparsity loss $\lambda_{sparse}=0.4$ that outperforms the best method by \textbf{7.21\%} using $\sim$\textbf{9.73} $\times$ fewer parameters.

\renewcommand{\arraystretch}{0.7}%
\begin{table}[!htb]
    \begin{subtable}[t]{.6\linewidth}
        
        \caption{Protocol 1: 5-layer CNN}
      \resizebox{\linewidth}{!}{
\begin{tabular}{c | c | c| c } 
 \hline 
 Method & ACC\% & BWT\% & Size(MB) \\ [0.3ex] 
 \hline\hline
 HAT \cite{Serr2018OvercomingCF} & $76.96(1.23)$ & $0.01(0.02)$ & $27.2$ \\ 
 PNN \cite{Rusu2016ProgressiveNN} & 75.25(0.04) & 0.00(0.00) & 93.51 \\

 A-GEM \cite{Chaudhry2019EfficientLL} & 54.38(3.84) & -21.99(4.05) & 25.4 \\
  ER-RES \cite{Chaudhry2019ContinualLW} & 66.78(0.48) & -15.01(1.11)  & 25.4 \\
 ACL \cite{ebrahimi2020adversarial} & 78.08(1.25) & 0.00(0.01) & 25.1 \\ [0.5ex]
 \hline\hline
 
 Baseline\textunderscore UB
 & 89.2(0.32) & 0.00(0.00) & 31.67 \\ 
 \textbf{CACL ($\lambda_{sparse}=0.4$)} & \textbf{83.71(0.19)} & \textbf{0.00(0.00)} & \textbf{2.58} \\ 
 \textbf{CACL ($\lambda_{sparse}=0.1$)} & \textbf{86.19(0.38)} & \textbf{0.00(0.00)} & \textbf{8.68} \\ 
 \hline
\end{tabular}\label{protocol1}
}
    \end{subtable}%
    \begin{subtable}[t]{.4\linewidth}
        \caption{Protocol 2: VGG\textunderscore16BN}
      \resizebox{\linewidth}{!}{
\begin{tabular}{c | c | c } 
 \hline
 Method & ACC\%  & Size(MB) \\ [0.3ex] 
 \hline\hline
 PackNet \cite{Mallya2018PackNetAM} & $67.5$ & $128.25$ \\ 
 PAE \cite{PAE} & $71.1$ & $256.5$ \\
 CPG \cite{hung2019compacting} & $80.9$ &$278$ \\ 
 \hline\hline
 \textbf{CACL} & \textbf{90.5}  & \textbf{73} \\ 
 \hline
\end{tabular}
}\label{protocol2}
    \end{subtable} \label{tab:cifar100}
    \caption{Results on 20-Split CIFAR-100 on (a) \emph{Protocol 1} and (b) \emph{Protocol 2}. (a) Comparison of CACL trained on 5-layer convolutional network with the \emph{state-of-the-art}. (b)  Comparison of CACL approach with \emph{network expansion} methods. CACL uses  VGG\textunderscore16BN ignoring the final linear layers while other works use the entire VGG\textunderscore16BN architecture.}
\end{table}

\textbf{Protocol 2}: In this protocol, we adopt the evaluation settings of a recent network based expansion method \cite{hung2019compacting}. Authors of \cite{hung2019compacting} employ VGG16 \cite{VGG} network with  batch normalization having \emph{separate} normalization parameters for each task. We refer to this network as VGG16\textunderscore BN. This protocol sheds light on the \emph{scalability} of our approach to other architectures. We compare our method with other network expansion methods in this protocol. We use a slightly different network architecture from \cite{hung2019compacting} by considering only the \emph{feature-extractor} of VGG16\textunderscore BN and ignore the final linear layers of the original VGG16\textunderscore BN network. We further maintain same number of task-head parameters as \cite{hung2019compacting} for fair comparison. Detailed architecture has been provided in the supplementary material. Results are reported in Table \ref{protocol2}. Despite starting with a smaller network (due to removal of final linear layers), our method \emph{remarkably} outperforms other network expansion based methods. Our best model with  $\lambda_{sparse}=0.01$ outperforms \cite{hung2019compacting} with \textbf{11.8}\%  improvement in accuracy
 on a model that is $\sim$ \textbf{3.8} $\times$ smaller in size. In fact, our method starts with an initial model of size 58.5MB and grows to the size 73MB after learning 20 tasks with an expansion rate of \textbf{0.24} when compared to \cite{hung2019compacting} that grows by \emph{1.14}. This shows that our method is not only scalable but also offers highly optimal and compression-friendly solution for incremental task learning when compared to other network expansion methods.

\subsection{miniImageNet}
\label{sec:miniImageNet}
We evaluate our appraoch on  miniImageNet\cite{miniImageNet} following the task splitting settings of \cite{ebrahimi2020adversarial, Chaudhry2019ContinualLW} for fair comparison. 
We split the dataset into 20 tasks with 5 classes per task. Table \ref{miniImageNet} demonstrates the performance of our method on this dataset. Results of the other works used for comparison are obtained from \cite{ebrahimi2020adversarial}.
Our method ($\lambda_{sparse}=0.1$) showcases  \textbf{12.3\%} improvement in accuracy over the best method \cite{ebrahimi2020adversarial} while using a model with $\sim$ \textbf{8.6} $\times$ fewer number of parameters. Further, our model trained with \emph{higher} sparsity loss ($\lambda_{sparse}=0.4$) outperforms the state-of-the art showing \textbf{9.3\%} improvement using $\sim$ \textbf{14} $\times$ fewer parameters.

In essence, our approach is composed of \emph{three} major building blocks namely, \emph{factorization with compression}, \emph{shared representations} and \emph{expansion}.  To illustrate the contribution of the individual components in the overall performance and compression achieved, we experiment by making minor alterations to our Algorithm \ref{CACL_algorithm}. We demonstrate the advantages of the proposed shared representations
under \emph{fixed} memory settings by training a network following Algorithm \ref{CACL_algorithm} and skipping the expansion step \emph{except} for task $t=1$. We refer to this scenario as \textbf{CACL-Fixed}. In another scenario, we train individual single task models that incorporates factorization and compression without sharing, referred to as \textbf{CACL-ST} (that is, we follow Algorithm \ref{CACL_algorithm} and ignore the final concatenation step). We compare the above models with \textbf{Baseline\_UB} and our final model (\textbf{CACL}) to elucidate the role of each component. Results are summarized in Table \ref{tab:ablations}. Firstly, we observe that CACL-ST when compared to Baseline\_UB significantly reduces the overall model size with minimal performance degradation. This justifies that \emph{factorization with compression} enables highly compact representations without the need of expensive \emph{re-training}. Secondly, with sharing enabled, our final CACL model performs competitively compared to CACL-ST while having a smaller model size. This shows that \emph{shared representations} contribute to model compression. Finally, CACL-Fixed when compared to CACL-ST provides competitive performances with very small model size. This suggests that our \emph{shared representations} can handle the problem of incremental task learning reasonably well under memory constrained settings. We report the results of such experiments on other datasets in the supplementary material.

 \begin{table}[!tb]
\vspace{-4mm}
    \begin{subtable}[t]{.6\linewidth}

        \caption{20-Split miniImageNet}
      \resizebox{\linewidth}{!}{
\begin{tabular}{c | c | c| c } 
 \hline
 Method & ACC\% & BWT\% & Size(MB) \\ [0.5ex] 
 \hline\hline
 HAT \cite{Serr2018OvercomingCF} & $59.45(0.05)$ & $-0.04(0.03)$ & $123.6$ \\ 
 PNN \cite{Rusu2016ProgressiveNN} & $58.96(3.50)$ & $0.00(0.00)$ & $588$ \\
 ER-RES \cite{Chaudhry2019ContinualLW} & 57.32(2.56) & -11.34(2.32) & 102.6 \\ 
 A-GEM \cite{Chaudhry2019EfficientLL} & 52.43(3.10) & -15.23(1.45)  & 102.6 \\
 ACL \cite{ebrahimi2020adversarial} & 62.07(0.51) & 0.00(0.01) & 113.1 \\ 
 \hline\hline
 Baseline\textunderscore UB
 & 73.6(0.08) & 0.00(0.00) & 37.8 \\ 
 
 \textbf{CACL ($\lambda_{sparse}=0.4$)} & \textbf{67.84(0.02)} & \textbf{0.00(0.00)} & \textbf{8.1} \\ 
 \textbf{CACL ($\lambda_{sparse}=0.1$)} & \textbf{69.67(0.77)} & \textbf{0.00(0.00)} & \textbf{13.1} \\ 
 \hline
\end{tabular}
}
\label{miniImageNet}
    \end{subtable}%
    \begin{subtable}[t]{.4\linewidth}

        \caption{Case Study}
      \resizebox{\linewidth}{!}{
\begin{tabular}{c | c | c } 
 \hline
 Method & ACC\%  & Size(MB) \\ [0.5ex] 
 \hline\hline
 Baseline\textunderscore UB & $73.6(0.08)$ & $37.8$ \\ 
CACL-ST  & 72.76(0.40) &  14.82 \\ 
 
 CACL-Fixed
 & 69.2(0.83)  & 8.10 \\ 
 CACL & 69.67(0.77)  & 13.14 \\ 

 \hline
\end{tabular}
}\label{tab:ablations}
    \end{subtable} 
    \caption{Results on 20-Split miniImageNet. (a) We compare our method with recent \emph{state-of-the-art} approaches using a 5-layer convolutional neural network. (b) Ablation study of our method on the miniImageNet dataset.}
\end{table}

\subsection{5-sequence dataset}
This dataset proposes sequential task learning on 5 different datasets in which each task learns a \emph{10}-way classifier on one of the datasets. The 5 datasets includes \textbf{SVHN}, \textbf{CIFAR10}, \textbf{not-MNIST}, \textbf{Fashion-MNIST} and \textbf{MNIST}.

\begin{figure}[h!]
\vspace{-2mm}
\begin{floatrow}
\ffigbox[0.78\FBwidth]{%
  \includegraphics[width=\linewidth]{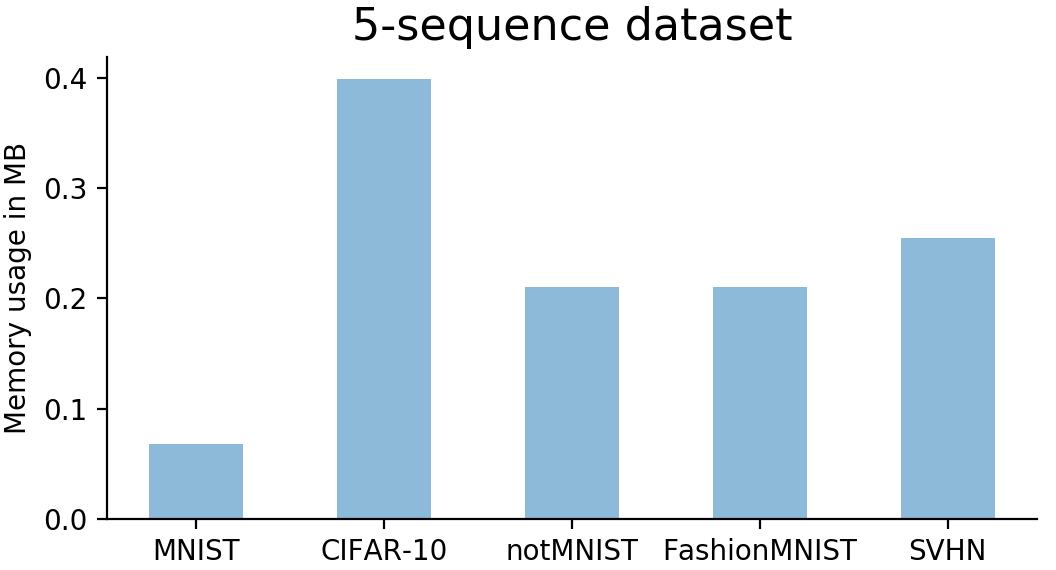}%
  
}{%
  
  \caption{Dynamic memory allocation on 5-sequence dataset}%
  \label{fig:memorydistrbution}
} 
\capbtabbox[1.2\FBwidth]{%
\resizebox{\linewidth}{!}{
\begin{tabular}{c | c | c| c } 
 \hline
 Method & ACC\% & BWT\% & Size(MB) \\ [0.6ex] 
 \hline\hline
 UCB \cite{Uncertainty} & $76.34(0.12)$ & $-1.34(0.04)$ & $32.8$ \\ 
 ACL \cite{ebrahimi2020adversarial} & 78.55(0.29) & -0.00(0.15) & 16.5 \\ 
 \hline\hline
 Baseline\textunderscore UB
 & 93.05(0.23) & 0.00(0.00) & 8.02 \\ 
 \textbf{CACL ($\lambda_{sparse}=0.1$)} & \textbf{90.84(0.13)} & \textbf{0.00(0.00)} & \textbf{1.35} \\ 
 \hline
\end{tabular}}
}
{%
  \caption{Comparison of CACL and\emph{state-of-the-art} approaches on 5-sequence dataset}%
  \label{5sequencedataset}
}
\end{floatrow}
\end{figure}

Results on this dataset are reported in Table \ref{5sequencedataset}.
Our method shows \emph{significant} improvements over the best method \cite{ebrahimi2020adversarial} yielding \textbf{15.6}\%  increase in accuracy  with $\sim$ \textbf{12.7}$\times$ fewer number parameters. An interesting observation is that the initial network (a 5-layer convolutional network) is of size 1.75MB while the final trained model obtained with CACL is of size \textbf{1.35}MB. This shows that our method produces models with minimal memory usage. In contrast to the recent works \cite{Rusu2016ProgressiveNN, ebrahimi2020adversarial} that propose static allocation of parameters for each task, our approach demonstrates dynamic resource allocation based on the task. To analyse the dynamic memory allocation in CACL, we show the individual memory usage by each task corresponding to a unique dataset in Fig. \ref{fig:memorydistrbution}. We observe that \emph{CIFAR-10} \cite{CIFAR100} consumes the highest memory of all tasks, while \emph{MNIST} the least. Also, \emph{not-MNIST} and \emph{Fashion-MNIST} are alloted similar number of parameters. Intuitively, we can gather from Fig. 2 that our CACL approach assigns dynamic memory based on the task difficulty which can be very beneficial for optimal memory/compute usage in real-world continual learning settings.

\section{Conclusion}
\label{conclusion}
In this work, we have proposed a compression based continual learning method that 
can dynamically grow a network to accommodate new tasks and uses low-rank approximations to achieve compact task representations. By encouraging the model to learn low-rank weight filters, we minimize the performance degradation during the compression phase and eliminate the \emph{time-consuming} retraining step.
By incorporating network factorization and a novel shared representational space, our method demonstrates \emph{state-of-the-art} results on three datasets with highly compressed models. Our method manifests scalability to other architectures and also exhibits dynamic resource allocation based on the task.
In this work, we assumed clear task boundaries exist between tasks and maintain individual task-identifiers during continual learning. In future work, we wish to eliminate such identifiers and further extend the compression/factorization strategy to linear layers.

\section*{Acknowledgement}
This work was supported in part by the National Science Foundation, and the Amazon Research Award.

\newpage
\section*{Supplementary Material}
\appendix
\renewcommand{\thesection}{\Alph{section}}

\section{Architectural Details}

\subsection{5-layer CNN}
\textbf{Architectural details} : We employ a simple 5 layer convolutional neural with convolutional layers of shape $64(3\times3) - 64(3\times3)- 128(3\times3) - 128(3\times3) - 256(2\times2)$, where each convolutional block is of shape $channel(kernel\_height \times kernel\_width)$. We introduce intermediate dropout layers to regularize training. For each incoming task $t$ we attach $T^t_{head}$, where $T^t_{head}$ consist of a \emph{single} linear layer that converts the convolutional features to task output. 

\textbf{Training details}: We used $Adam$ optimizer for all the experiments with base learning rate $1e^{-3}$. We train each task for \emph{200} epochs and drop the learning rate by factor of 10 at \emph{80, 120, 180} epochs. We set $\lambda_{orth}=1.00$ to ensure equal importance to orthogonality regularizer($L_{orth}$) as task specific loss($L_{task}$). We set the pruning intensity parameter e to $1e^{-5}$.

\subsection{VGG16\textunderscore BN}
\textbf{Architectural details} : We use the \emph{feature-extractor} of VGG16\textunderscore BN ignoring the final linear layers \emph{fc6, fc7, fc8} of VGG16 \cite{VGG}. We use two linear layers for each task \emph{t} as task-head $T^t_{head}$ to maintain similar number of task-specific parameters as in \cite{hung2019compacting}. The four batch normalization parameters namely \emph{mean, variance, running\textunderscore mean, running\textunderscore variance} for each batch-norm layer are stored separately for each task $t$ similar to \cite{hung2019compacting}.

\textbf{Training details}: We followed the same training strategy as for the 5-layer CNN. However, we set the sparsity loss weight,  $\lambda_{sparse}$, to be $0.01$. 

\section{Algorithms}
We provide the pseudo code for our compression-aware training(Section \ref{training}) and compression(Section \ref{compression}) in Algorithm \ref{train_algorithm}, Algorithm\ref{compression_algorithm} respectively. 

    \begin{minipage}[t]{\textwidth}
\begin{algorithm}[H]
\small
\SetAlgoLined
  \SetKwFunction{FTN}{TrainTask}
  \SetKwProg{Fn}{function}{:}{}
  \Fn{\FTN{ $D_{t}$, \:  $[\mathcal{U}^{\leq t-1}, \mathcal{S}^{\leq t-1}, \mathcal{V}^{\leq t-1} ]$,   [ $\mathcal{U}^{t}, \mathcal{S}^{t}, \mathcal{V}^{t}$  ]}}{

        \For{$epoch\gets1, 2, ... epochs$}{
        \For{$batch\gets1, 2, ... batches$}{
        ($x_{batch}, y_{batch}$) $\sim$  $D_t$ \\

        $\mathcal{U}^{t}, \mathcal{V}^{t} \gets \mathcal{U}^{t} - \nabla_{\mathcal{U}^t} (L_{task} + \lambda_{orth}L_{orth} ), \mathcal{V}^{t} - \nabla_{\mathcal{V}^t}(L_{task} + \lambda_{orth}L_{orth})$ \\
        
        $\mathcal{S}^t \gets \mathcal{S}^t + \nabla_{\mathcal{S}^t} (L_{task} + \lambda_{sparse}L_{sparse} )$

         }
         
        }
        \KwRet $\mathcal{U}^{t},\mathcal{S}^t, \mathcal{V}^{t}$\\
  }
 \caption{Compression-aware training on task \emph{t}}
 \label{train_algorithm}
\end{algorithm}
\end{minipage}

    \begin{minipage}[t]{\textwidth}
\begin{algorithm}[H]
\small
\SetAlgoLined
  \SetKwFunction{FCND}{Compression}
  \SetKwProg{Fn}{function}{:}{}
  \Fn{\FCND{ [$\mathcal{U}^{t},\mathcal{S}^{t}, \mathcal{V}^{t}$], $e$}}{

        \For{$l\gets1, 2, ... layers$}{
        
         $U, V, S = U^t_l, V^t_l, S^t_l$  
          \Comment*[r]{$U \in R^{c\times r}, V \in R^{nhw\times r}$} 
          
        $ind = argsort(\{\sigma^t_{i\; l}\}^r_{i=1}, desc)$ 
        \Comment*[r]{Singular value indices \emph{descending} order} 
        $U_{sort}, V_{sort}, S_{sort} = \; U[ind], V[ind], S[ind]$ \Comment*[r]{Sort the vectors}
        
        $topk, energy_{curr}  = \; 0, 0$ \\
        
        $energy_{tot}  =  \sum_{i=1}^r (\sigma_{i\; l}^t)^2$ \Comment*[r]{Total singular energy}        
        \texttt{\\}

        \While{$\frac{energy_{curr}}{energy_{tot}}  < 1 - e$}
        {
             \texttt{\\}
          $energy_{curr}$ += $ (\sigma^t_{topk\; l})^2$ \\
            topk = topk + 1 
         }

        $U^t_{pruned \: l}, V^t_{pruned \: l}, S^t_{pruned \: l} \gets U_{sort}[:topk], V_{sort}[:topk], S_{sort}[:topk] $ \\ 
        
        \Comment*[r]{$U^t_{pruned \: l} \in R^{c\times topk}, V^t_{pruned \: l} \in R^{nhw\times topk} $}
        }
        \KwRet $\mathcal{U}^{t}_{pruned},\mathcal{S}^t_{pruned},\mathcal{V}^{t}_{pruned}$
  }
 \caption{Compression for task $t$}
 \label{compression_algorithm}
\end{algorithm}
\end{minipage}

\section{Additional Experiments}
We present case-study results using CIFAR100 dataset(results using miniImageNet are discussed in  Section \ref{sec:miniImageNet} in the main paper) in Table \ref{Case Study on CIFAR100}. We find that CACL-ST effectively performs model compaction with minimal accuracy degradation. However, we find that CACL with shared representations consumes more memory than CACL-ST. This suggests selective sharing of representations might better help and we wish to explore this in our future work. Finally, we find that CACL-Fixed with limited memory constraint performs competitively when compared to others. 

 \begin{table}[!hb]

\begin{tabular}{c | c | c } 
 \hline
 Method & ACC\%  & Size(MB) \\ [0.5ex] 
 \hline\hline
 Baseline\textunderscore UB &  $89.2(0.32)$ & $31.67$ \\ 
CACL-ST  & 88.9(0.12) &  7.57 \\ 
 
 CACL-Fixed
 & 82.15(1.07)  & 1.96 \\ 
 CACL & 86.19(0.38)  & 8.68 \\ 

 \hline
\end{tabular}
\caption{Case Study on CIFAR100}
\label{Case Study on CIFAR100}
\end{table}

\end{document}